\documentclass[aip,jcp,preprint,amsmath,amssymb]{revtex4-1}

\usepackage{dcolumn}
\usepackage{bm}
\usepackage{upgreek}
\usepackage{array}
\usepackage{geometry}
\usepackage{siunitx}
\usepackage{hyperref}
\usepackage{graphicx}
\usepackage{tabularx}
\usepackage{booktabs}
\usepackage{url}
\usepackage{hyperref}
\usepackage{natbib}
\usepackage{xcolor}
\hypersetup{colorlinks, citecolor={blue}, linkcolor={red}, urlcolor={violet}, pdftitle={Machine Learning and VIIRS Satellite Retrievals for Skillful Fuel Moisture Content Monitoring in Wildfire Management}, pdfauthor={J. S. Schreck et al.}, pdfdisplaydoctitle}

\begin{document}
\title{Machine Learning and VIIRS Satellite Retrievals for Skillful Fuel Moisture Content Monitoring in Wildfire Management}

\author{John S. Schreck}
\email{schreck@ucar.edu}
\affiliation{National Center for Atmospheric Research (NCAR), Computational and Information Systems Laboratory, Boulder, CO, USA}

\author{William Petzke}
\affiliation{National Center for Atmospheric Research (NCAR), Research Applications Laboratory, Boulder, CO, USA}

\author{Pedro A. Jim\'enez}
\email{jimenez@ucar.edu}
\affiliation{National Center for Atmospheric Research (NCAR), Research Applications Laboratory, Boulder, CO, USA}

\author{Thomas Brummet}
\affiliation{National Center for Atmospheric Research (NCAR), Research Applications Laboratory, Boulder, CO, USA}

\author{Jason C. Knievel}
\affiliation{National Center for Atmospheric Research (NCAR), Research Applications Laboratory, Boulder, CO, USA}

\author{Eric James}
\affiliation{Cooperative Institute for Research in Environmental Sciences (CIRES), University of Colorado, Boulder, CO, USA}
\affiliation{Global Systems Laboratory, National Oceanic and Atmospheric Administration (NOAA), Boulder, CO, USA}

\author{Branko Kosovi\'c}
\affiliation{National Center for Atmospheric Research (NCAR), Research Applications Laboratory, Boulder, CO, USA}

\author{David John Gagne}
\affiliation{National Center for Atmospheric Research (NCAR), Computational and Information Systems Laboratory, Boulder, CO, USA}

\begin{abstract}
Monitoring the fuel moisture content (FMC) of vegetation is crucial for managing and mitigating the impact of wildland fires. The combination of in situ FMC observations with numerical weather prediction (NWP) models and satellite retrievals has enabled the development of machine learning (ML) models to estimate dead FMC retrievals over the contiguous US (CONUS). In this study, ML models were trained using variables from the National Water Model and the High-Resolution Rapid Refresh (HRRR) NWP models, and static variables characterizing the surface properties, as well as surface reflectances and land surface temperature (LST) retrievals from the VIIRS instrument on board the Suomi-NPP satellite system. Extensive hyper-parameter optimization yielded skillful FMC models compared to a daily climatography RMSE (+44\%) and to an hourly climatography RMSE (+24\%). Furthermore, VIIRS retrievals were important predictors for estimating FMC, contributing significantly as a group due to their high band-correlation. In contrast, individual predictors in the HRRR group had relatively high importance according to the explainability techniques used. When both HRRR and VIIRS retrievals were not used as model inputs, the performance dropped significantly. If VIIRS retrievals were not used, the RMSE performance was worse. This highlights the importance of VIIRS retrievals in modeling FMC, which yielded better models compared to MODIS. Overall, the importance of the VIIRS group of predictors corroborates the dynamic relationship between the 10-h fuel and the atmosphere and soil moisture. These findings emphasize the significance of selecting appropriate data sources for predicting FMC with ML models, with VIIRS retrievals and selected HRRR variables being critical components in producing skillful FMC estimates.
\end{abstract}

\maketitle

\section{Introduction}

Wildland fires continue to have a significant impact on personal health/safety, the economy, infrastructure, and the environment. In the United States, the size and severity of fires have trended upwards over the past 30 years largely due to the effects of increased fuel loads from fire suppression, warmer and dryer climatic conditions, and the growth of human development in the Wildland Urban Interface (WUI) \cite{CBO}. One recent example of a particularly destructive wildfire is the Marshall Fire, which swept through the cities of Superior and Louisville, Colorado on December 30, 2021. Dry, windy conditions contributed to one or more grass fires spreading rapidly into urban areas. The fire destroyed 1,084 buildings, led to two fatalities, and had an estimated cost in excess of \$500 million \cite{wiki:marshall}. Improved fire modeling will provide land managers and public safety officials with better situational awareness of changing fire risk and help predict the spread of dangerous wildfires such as the Marshall Fire.

One of the most important factors in improving current numerical fire modeling, for example with WRF-Fire \cite{coen2013wrf}, is providing accurate estimates of current fuel moisture as inputs to the model. Rothermel \cite{rothermel1972} discusses the impact of fuel moisture on the completeness and rate of fuel consumption (reaction velocity), which is an important factor in model performance. Coen et al. \cite{coen2013wrf} performed sensitivity experiments on the WRF-Fire model that show significant changes in spread as fuel moisture values are increased or decreased. In general, increases in fuel moisture reduce fire spread and eventually approach or reach a point of extinction. One problem with the dependence of fire models on fuel moisture is the difficulty of finding high-quality gridded FMC data that covers the CONUS and Alaska (AK) at resolution required for effective fire spread prediction. Developing solutions to this problem is a primary objective of our research.

The two major categories of fuel moisture are live and dead fuel moisture \cite{yebra2013global}. Dead fuel moisture content (DFMC) is a key component of determining fire risk, and is dependent on weather conditions instead of other factors such as evapotranspiration \cite{wfas}. The Wildland Fire Assessment System (WFAS) \cite{wfas} currently provides interpolated DFMC observations and forecasts for the CONUS and AK interpolated from in~situ observations at remote automated weather stations (RAWS). The observation data from these stations are used as predictand values in this research. 

Numerous studies have noted the effectiveness of using meteorological observations for retrieving DFMC estimates \cite{chuvieco2004conversion, aguado2007estimation, nolan2016predicting, nolan2016large, boer2017changing, hiers2019fine, lee2020estimation, cawson2020corrigendum, masinda2021prediction}. However, RAWs are relatively sparsely distributed and can lead to complications in operational deployment of FMC estimation products \cite{rs13214224}. More recently, remote sensing (satellite) retrievals, in particular MODIS and MSG-SEVIRI, have been used to overcome limitations from other ground observations for retrievals of both live \cite{marino2020investigating,yebra2013global,caccamo2011monitoring,stow2006time,peterson2008mapping} and dead FMC \cite{NIETO2010861,nolan2016predicting,zormpas2017dead,McCandless2020,rs13214224}. MODIS provides different spatial resolutions while MSG-SEVIRI provides high temporal resolution. For example, Nieto et~al.\ used MSG-SEVIRI retrievals to provide hourly estimates of equilibrium moisture content. More recently, Dragozi et al.\ showed that MODIS reflectance bands provided satisfactory accuracy in DFMC estimation for wildfires in Greece \cite{rs13214224}. 

In this work, we investigate the effectiveness of estimating DFMC using the satellite reflectance bands from the Visible Infrared Imaging Radiometer Suite (VIIRS) instrument. VIIRS is seen as a replacement for the MODIS platform on the Terra and Aqua satellites, which have data degradation issues including band striping/noise and are approaching end of life. The data obtained from VIIRS have higher spatial resolution than MODIS (375m vs. 1km), which is clearly important for fire modeling applications. MODIS is also limited to CONUS while VIIRS captures CONUS along with Alaska. VIIRS, like MODIS, provides data twice a day. 

We also focus on using machine learning as the primary modeling approach for using meteorological and remote sensing observations to make DFMC predictions at sites spread across CONUS. The use of machine learning for the prediction of FMC has been growing in recent years \cite{McCandless2020, fan2021physics, SHMUEL2022119897, xie2022retrieval, fan2021physics, capps2021modelling, zhu2021live}. In particular, this work builds on that performed by McCandless et al. \cite{McCandless2020}, which utilized MODIS reflectance bands and National Water Model (NWM) data paired with RAWS data in Random Forest (RF) and neural network (NN) models to predict DFMC for CONUS. Other recent advancements include the application of support vector machines (SVM) \cite{lee2020estimation}, convolutional neural networks \cite{zhu2021live} and long-short term memory (LSTM) networks \cite{fan2021physics}. As described in Section \ref{datasets}, the data sets that are used to train machine learning models are tabular, hence only linear regression, gradient boosting approaches, and standard feed-forward, fully connected neural networks are considered herein, as more complex ML approaches have not been found to produce better performance. Expanding on McCandles’ application, several explainability methods are herein applied to the ML models as a means for identifying the most important predictors. We also probe the most important predictors by group (e.g., VIIRS, weather inputs, etc.). These last two investigations are important for identifying whether the ML predictions make physical sense, as well as for designing a minimal model for use in operation. 

Our ML approach uses satellite reflectance bands from the VIIRS instrument and weather model data such as the High-Resolution Rapid Refresh (HRRR) to improve the estimation of DFMC at locations without nearby RAWS observations. This research builds on prior research done by McCandless et al. \cite{McCandless2020} which utilized MODIS reflectance bands and National Water Model (NWM) data paired with RAWS data in Random Forest (RF) models to predict DFMC for the CONUS. The Visible Infrared Imaging Radiometer Suite (VIIRS) instrument data has higher spatial resolution than MODIS (375m vs. 1km). This is important for improving fire modeling applications. VIIRS is seen as a replacement for the MODIS platform on the Terra and Aqua satellites which have data degradation issues including band striping/noise and are approaching end of life. In addition to new input data sets, our current research focuses on DFMC prediction with more advanced ML techniques including gradient-boosted trees and neural networks. We focus on performance optimization and model interpretability in depth.

One potential downside in the study by McCandles is that the performance of the trained models used are likely over-optimistic due the random splitting of the data used to perform cross validation \cite{McCandless2020}. They first split the data using 25 randomly chosen days held out as the test set. The remaining days were split randomly on site locations into training and validation sets (80/20). Random splitting essentially ignores specific space and time correlations that exist in fuel-moisture content training data sets, such as that used in this study. Therefore, models to predict FMC and trained on random splits likely represent (overly) optimistic performance, due to over-fitting on data present in hold-out splits that closely resemble examples in the training split. A primary future objective of this study is the application of models trained on CONUS to Alaska (and potentially Canada). As VIIRS has not been in operation for nearly as long as MODIS, the data sets cover a relatively short time period (three years). Hence, sites are withheld from the training data and are only present in either a test or validation split. This way of splitting aims to break the space and time correlations across the splits. Models trained on these data therefore should produce more realistic performance estimates.  

\section{Data sets}
\label{datasets}

The data set used to train and evaluate the machine learning models span a three-year period (2019--2021). The following sections describe the FMC observations used in the predictand data set (Section \ref{sec:predictands}), the predictor data sets (Section \ref{sec:predictors}), a correlation analysis of the predictand/predictor variables (Section \ref{sec:corr}), and how the training data set was split to independently train and validate the models (Section \ref{sec:validation}).

\subsection{Predictand data set\label{sec:predictands}}

In order to create the 10-h FMC data set, raw FMC observations were downloaded from the Meteorological Assimilation Data Ingest System (MADIS) archive (ftp://madis-data.ncep.noaa.gov//archive/) from 1 July 2001 through 31 December 2021. This archive contains hourly compressed NetCDF files and was 2.5T in total size.  After the full archive was downloaded, fuel moisture data and site information were extracted from the hourly files and combined into yearly NetCDF files. Only sites over CONUS were kept. During this phase, sites were removed that had either missing site identifiers, or inconsistent site location information. Many sites were reported with different location information throughout the year, so the site with the most recent location information was kept (i.e., site S with location X reported on 1 January 2010 would be removed if site S was reported with a different location any time after 1 January 2010). A data log containing sites that changed locations was maintained throughout the entire process. After the yearly files were made, the data from all years were combined into one NetCDF file containing sites over the CONUS region. Only sites which reported non-missing data between 2019--2021 were included in this file. In the end, there were 1823 RAWS sites. A QC flag was added to these files to indicate whether or not the data had passed a simple range check (0\%--400\%).  

Two separate climatographies were made from the FMC data using only data prior to 2019, one using only the Julian day (DOY) and another using Julian day along with the hour of the day (DOY-HR). In order to calculate the DOY climatography, each site’s data were combined over a 31-day window (15 days prior to the current day and 15 days after the current day) for all years making up the data set. For example, the climatography for site S on Julian day 145 would combine all site S data from 2001--2018 for Julian days 130--160. For Julian day 1, the climatography would consist of Julian days 351--16 (with 365 being 31 December). For leap years, data on February $29^{th}$ was combined with data on the $28^{th}$. After the data were combined for each site and day of year, the average, standard deviation, and count of the total data set were recorded. A minimum of six years of data was required for each day of the year. Otherwise, the average and standard deviation were set to missing. The DOY-HR climatography was calculated similarly to the DOY climatography, except that the data were further split on the hour of the day.

\subsection{Predictor data set \label{sec:predictors}}
The predictor data sets consist of variables from four different sources: static variables characterizing the surface characteristics including monthly climatographies from the Weather Research and Forecasting (WRF) Preprocessing System (WPS); analysis variables characterizing the near surface atmospheric conditions and the soil state from the High-Resolution Rapid Refresh (HRRR) model; hydrologic variables from the analysis of the National Water model (NWM); and surface reflectands (sfc rfl) retrievals (VNP09-NRT) and land surface temperature (LST) retrievals (VNP21-NRT) from the VIIRS instrument on board Suomi-NPP. The complete list of variables is shown in Table \ref{table:predictors}.

\begin{table*}
\centering
\begin{tabular}{ l | c | c | c }
\hline
Static & HRRR & NWM & VIIRS \\
\hline
Canopy fraction & 2-m temperature & Soil moisture & Sfc rfl M1 \\
Soil clay fraction & 2-m relative humidity & Evapotranspiration & Sfc rfl M2 \\
Urban fraction & Soil moisture availability & & Sfc rfl M3 \\
Elevation & Skin temperature & & Sfc rfl M4 \\
Impermeability & Mean sea level pressure & & Sfc rfl M5 \\
Irrigation & Canopy water & & Sfc rfl M7 \\
Land use & Snow cover & & Sfc rfl M8 \\
Soil sand fraction & Snow depth & & Sfc rfl M10 \\
Lowest soil category & 2-m dew point temperature & & Sfc rfl M11 \\
Top soil category & 2-m specific humidity & & Sfc rfl I1 \\
Snow albedo & 2-m potential temperature & & Sfc rfl I2 \\
Albedo clim & Cloud cover & & Sfc rfl I3 \\
Green fraction clim & Water equivalent snow depth & & LST \\
Leaf area index clim & Global horizontal irradiance & & \\
& Sensible heat & & \\
& Latent heat & & \\
& Ground heat & & \\
& Precipitable water & & \\
& Precipitation & & \\
& Precipitation rate & & \\
\hline\hline
\end{tabular}
\caption{\label{table:predictors}Variables from each predictor data set.}
\end{table*}

The HRRR (Dowell et al. 2021) is an operational hourly-updating numerical weather prediction model covering the CONUS with 3 km grid spacing.  The HRRR uses the Rapid Update Cycle (RUC) land surface model to represent the flow of moisture and energy between the atmosphere and land surface, with nine soil levels (Smirnova et al. 2016).  Importantly, the training and evaluation period of this study (2019-2021) spans two operational versions of the HRRR, HRRRv3 and HRRRv4.  More details on the HRRR configuration and performance differences by version are provided by Dowell et al. (2021) and James et al. (2021).  

The NWM is an operational hydrologic model covering the CONUS at 1 km grid spacing.  NWM receives its precipitation input from a variety of sources including quantitative precipitation estimates and model forecasts, the latter of which include HRRR for the short-range predictions (out to 18 h lead time).  

The predictor data sets have different spatio-temporal resolution, so some data manipulation was necessary to pair them with the predictand data set (see Section \ref{sec:predictands}) to create the training data set. The temporal pairing of HRRR and NWM variables is straightforward since they are both available every hour, which is the resolution of the predictand data set. Some manipulation was required to pair the monthly climatographies and VIIRS data sets. The climatographies were linearly interpolated to each day of the year, whereas VIIRS retrievals were assigned to the nearest hour. The VIIRS retrievals are available every 6 minutes, and each one of these granules were assigned to the nearest hour.

All the data sets were spatially interpolated into a grid over CONUS at 375-m grid spacing. This is the grid spacing of the finest resolution VIIRS channels (I bands). The NWM model grid spacing is 1 km. It was interpolated to the 375-m grid following a nearest neighbor interpolation. The nearest neighbor interpolation is the same approach used to interpolate the HRRR variables at 3 km grid spacing into the target grid of 375 m. In the case of VIIRS reflectances retrievals, only those retrievals without clouds or snow were interpolated into the 375-m grid. Again the nearest neighbor interpolation was used. This is the same interpolation procedure used for the land surface temperature retrievals available at 750-m grid spacing. Only the retrievals labelled as high or medium quality were used. Finally, the monthly climatographies available at 1-km grid spacing were interpolated to the 375-m grid following the nearest neighbor approach as well.

Hence, the majority of the predictors are available at a coarser grid spacing than the target grid at 375 m. To illustrate sensitivities to the target resolution, we also interpolated the predictors to a 2250-m grid spacing resolution using averages of the available data within the 2250-m grid cells.

In total, 44 million data points are associated with the predictors listed in Table \ref{table:predictors}. The four predictor groups do not have equivalent temporal spacing. For example, the VIIRS are only collected twice a day, meaning that for any one predictant (fuel moisture) value, not all predictors across the four groups will have values associated to them. We do not consider any `imputation' or other strategies for filling in missing values, so the choice of which predictors are selected as model inputs will determine the total amount of data where all predictor fields have finite values, and will influence a model's prediction performance. 

\subsection{Predictor/predictand correlations\label{sec:corr}}

\begin{figure}[t!]
    \centering
    \includegraphics[width=\columnwidth]{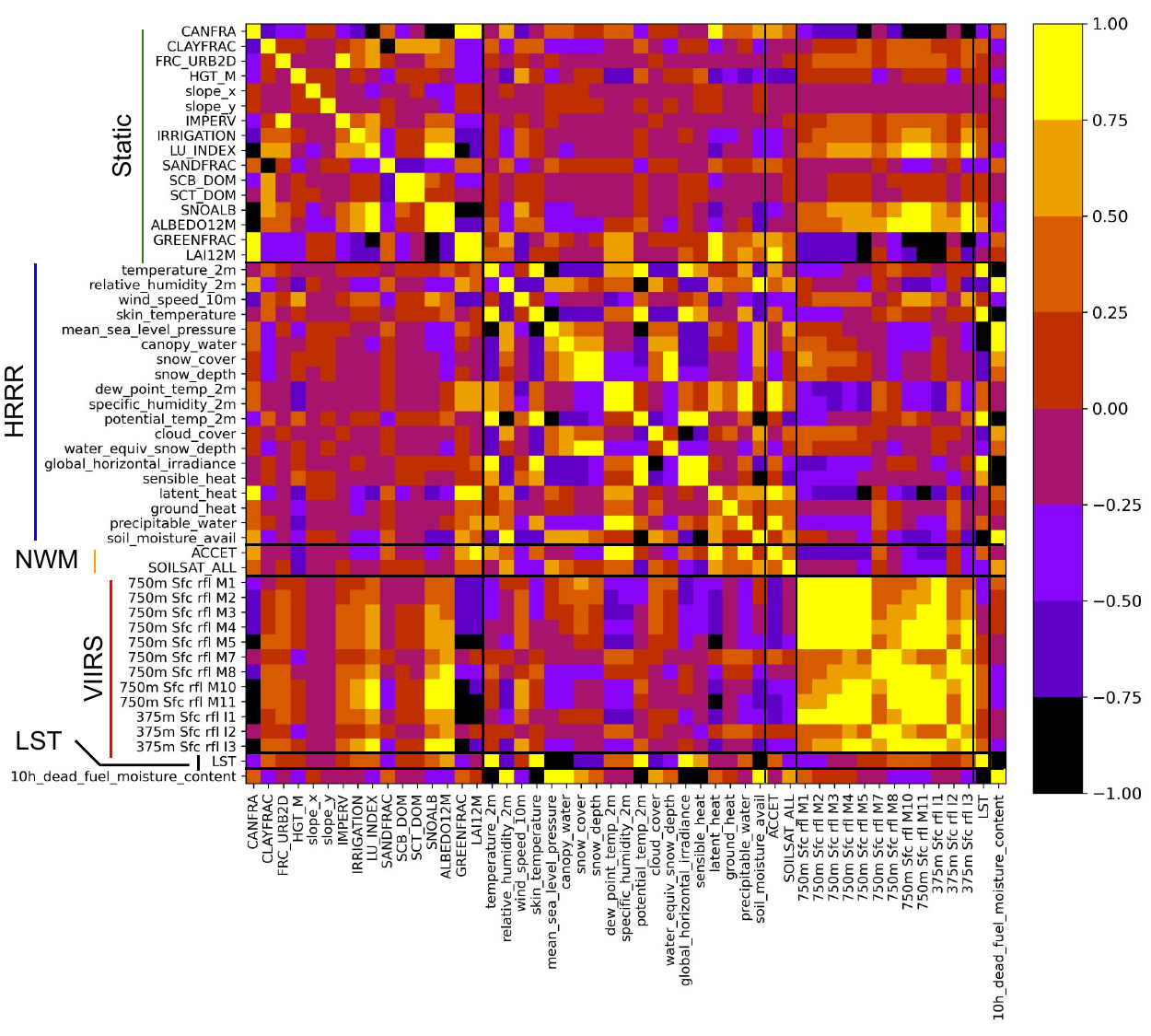}
    \caption{The computed correlation matrix is plotted for all of the predictors in Table~\ref{table:predictors} and the 10-h dead fuel-moisture content.}
    \label{fig:correlation}
\end{figure}

Selecting all predictors listed in Table~\ref{table:predictors}, as well as the 10-h DFMC (51 in total),  there are 940,000 data points where all 51 fields have finite values. Figure~\ref{fig:correlation} shows how the predictors from all of the groups correlate with each other and with the fuel moisture values. The figure shows that there are several HRRR predictors that are highly-correlated with the fuel moisture (both positive and negative), in particular, the temperature and water-associated variables. The HRRR temperature predictors also positively correlate strongly with the LST predictor, which also has high (negative) correlation with the DFMC. The LST predictor is also modestly correlated with the M10 and M11 VIIRS bands. 

Next, correlations are high among the VIIRS bands, but not any one band strongly correlates with DFMC with the exception of M10 and M11, for which the correlation is modest. The static variables including green fraction and albedo monthly climatographies are also observed to correlate strongly with the M8, M10, and M11 VIIRS band. However, only these and a few of the other static predictors show appreciable correlation with DFMC, and none of the NWM variables correlates strongly with DFMC. 

\subsection{Data splitting and standardization}
\label{sec:validation}

Once a selection of input groups is selected, the resulting data set is split into training (80\%), validation (10\%), and testing (10\%) data splits in order to train and test a ML model. Then, this is repeated 10 times via cross-validation by resampling the training and validation sets while holding the test set constant. As noted in the introduction, we consider two approaches to splitting: (1) by random selection, and (2) by randomly holding out sites (defined by latitude/longitude, of which there are about 1600 prior to January 1, 2019). The former selection does not consider that subsets of the data are highly correlated in either time, space, or both, therefore data points that are similar may end up in both training and validation/testing splits. In the latter approach, holding out a random subset of sites from a split aims to separate out those correlated data points correctly (e.g., they all should go to the same split). In both cases, a stratified approach was also used so that all three splits effectively had a representative sample of the FMC values. 

The relative ranges of all the predictors listed in Table \ref{table:predictors} need to be transformed into a new coordinate system so that the features having the largest spread do not dominate the weight space of an ML model. During model optimization (discussed below), we found that performance was usually better when the values for each quantity listed in Table \ref{table:predictors} and the fuel-moisture value were standardized independently into z-scores according to the formula $X_j = (X_j - u) / s$, wherein $u$ and $s$ are the mean and standard deviation of $X_j$, so that the mean is zero and the standard deviation is one (computed on the training set and then applied to the validation and test sets). The predicted FMC value is then inverse-transformed back into the original range observed in the training data set.

\section{Methods}

\subsection{Machine learning models}

\begin{figure}[t!]
    \centering
    \includegraphics[width=\columnwidth]{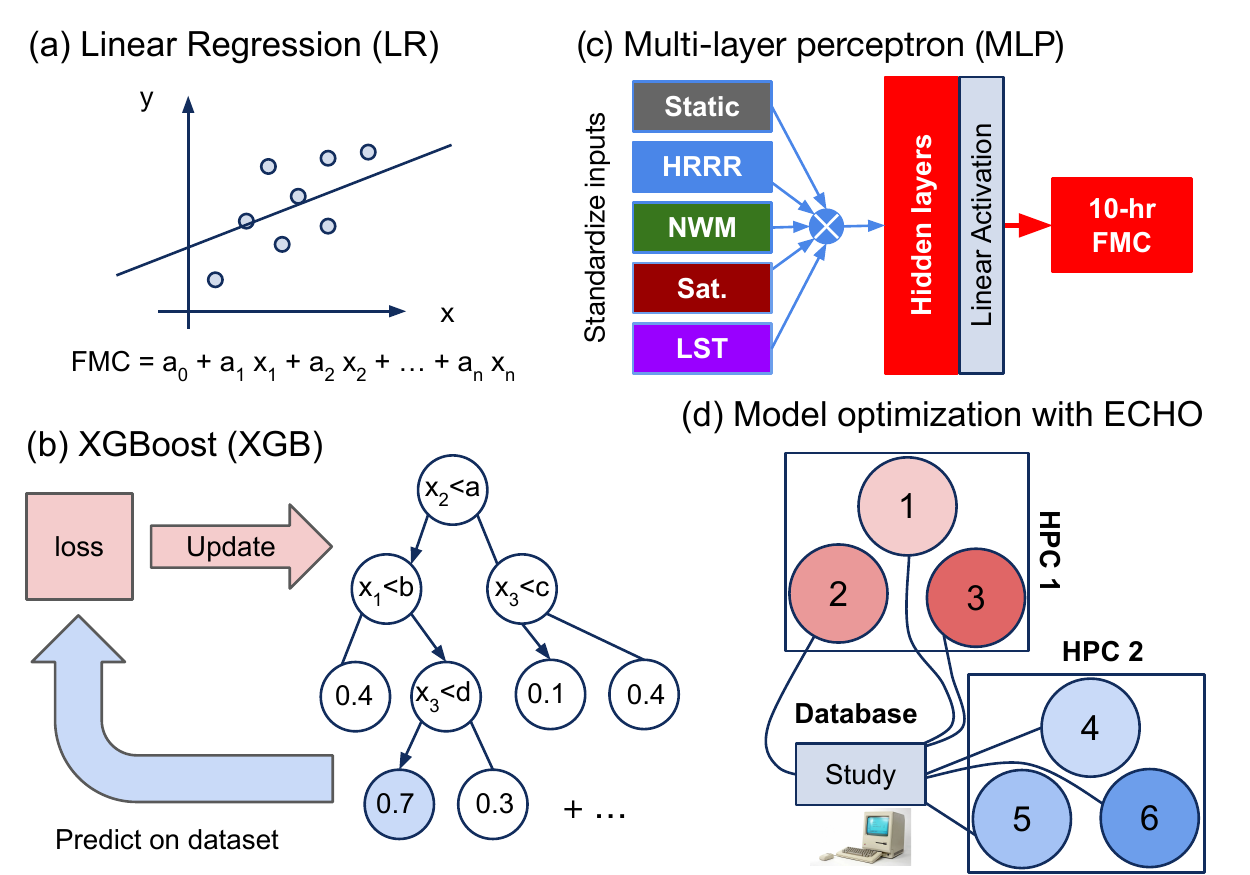}
    \caption{Schematic illustration of the three ML models considered: (a) Linear regression (LR), (b) XGBoost (XGB), and a feed-forward, multi-layer perceptron (MLP). (d) Illustration of the scalable hyperparameter optimization approach used by ECHO to find the best performing XGB and MLP models.}
    \label{fig:models}
\end{figure}

Figure \ref{fig:models}(a) illustrates three ML models considered: (a) linear regression (LG), (b) a scalable, distributed gradient-boosted decision tree (Extreme Gradient Boosting, e.g., the XGBoost ``model'') \cite{xgboost}, and (c) a vanilla feed-forward multi-layer perceptron (MLP) similar to those used in \cite{mlp_model}. The LR approach was chosen as the baseline ML model because it assumes a linear dependence of the FMC on each predictor, for $n$ total predictors, and will be the simplest model considered.

The XGB and MLP models are both non-linear and of increasing model and training complexity, both relative to LG, and often MLP relative to XGB. The XGB model was chosen because it usually outperforms deep learning methods for tabular data sets \cite{kossen2021self}, and thus is commonly recommended \cite{tabular_xgb1, tabular_xgb2}. XGB also typically requires much less tuning compared to MLPs, and is usually much faster to use.  These are important considerations because the model is to be deployed operationally in the near future. However, neural network ensembles are often as performant as XGB, and there are more methods available to probe model explainability and uncertainties associated with model predictions, so we investigate them as well. 

With the linear regression model, the values of coefficients $a_i$ in the figure corresponding with the input features $x_i$ can be computed from $n+1$ equations using least squares. The number of data points is much larger than $n$ and thus we have a statistically robust determination of the coefficients $a_i$. XGBoost uses a weighted sum from an ensemble of decision trees to make FMC predictions (a single tree is shown in the Figure \ref{fig:models}(b)). Each iteration in the algorithm uses the computed root-mean-square error (RMSE) from the previous round to make the current fit. We initially also considered random forests but found them always to be inferior to XGBoost, so we only focused on the XGBoost model.

The MLP comprises a stack of $N$ fully-connected, feed-forward layers and activation functions. The first input layer transforms the chosen input groups into a representation of size $L$ and is followed by a LeakyReLU activation. The last layer transforms the latent representation of the input described by preceding layers into size one and uses a linear activation. The neural network becomes deep if there are subsequent layers sitting between the first and last layers of size $L$, and separated from each other by LeakyReLU activation functions. After all LeakyReLU activation functions, we used a 1D batch normalization and a dropout layer, respectively. 

For the MLP, the weights of the model are updated by computing a training loss on a batch of inputs and then using gradient descent with back-propagation \cite{rumelhart1986}, along with a pre-specified learning rate to reduce the training error. Model training involves repeating this process and occasionally computing the RMSE on the validation split, then stopping training once the validation RMSE is no longer improving. We reduced the learning rate by a factor of ten when the validation loss reached a plateau. 

The XGB and MLP models were both subject to extensive hyperparameter optimization to find the best performing models according to the RMSE computed on the validation split. We used the Earth Computing Hyperparameter Optimization package to perform a scaled optimization for both models, performing 1,000 hyper parameter selection trials for XGB and MLP models, for random and site splits (at 375-m and 2250-m), as is illustrated in Figure \ref{fig:models}(d). The first 100 trials used random parameter sampling and then from 100 onward a Bayesian approach using a Gaussian mixture model strategy was employed to perform an informed search (see \cite{bergstra2011algorithms} for more details). For XGB, the varied hyperparameters were the learning rate, the minimum drop in loss needed to make a further partition on a leaf node of the tree (denoted $\gamma$), the maximum tree depth, the number of estimators (boosting rounds), the training sub-sample rate, and the sub-sample ratio of columns when constructing each tree. For the MLPs, the varied parameters were the number of layers $N$, the size of each layer $L$, the learning rate, training batch size, the L2 penalty, and the selection of the training loss. The training loss choices were the mean-absolute error, the RMSE, the Huber loss, or log-hyperbolic cosine (log-cosh) loss.
  
\subsection{Metrics}

As noted, the RMSE was used as the validation and testing metric for all three models. In addition to the RMSE, the coefficient of determination is also used as a performance metric. They are defined as
\begin{eqnarray}
    RMSE &= \left(\frac{1}{N} \sum_{i=1}^{N} (y_i - f(x_i))^2\right)^{1/2} \\ 
    R^2 &= 1 - \frac{\sum_{i=1}^{N} (y_i - f(x_i))^2}{\sum_{i=1}^{N} (y_i - \bar{y})^2},
\end{eqnarray}
wherein $x_i$ represents the $ith$ vector input of predictors, $f(x_i)$ is the predicted $ith$ fuel moisture value, $y_i$ is the $ith$ true fuel moisture value, $\bar{y}$ is the average fuel moisture for the data set, and $N$ is the data set size. To determine how well the model performance compares relative to climatographies for the DFMC, two skill scores are defined as 
\begin{eqnarray}
    \label{skill_rmse}
    Skill(RMSE) &= 1 - \frac{RMSE(ML)}{RMSE(Clim)} \\ 
    \label{skill_r2}
    Skill(R^2) &= \frac{R^2(ML) - R^2(Clim)}{1 - R^2(Clim)}.
\end{eqnarray}
A skill score larger than zero means the model outperforms the climatography on this metric (1 indicates the model is perfect), while less than zero means the climatography estimate is better (zero means the model and the climatography are equivalent in terms of metric performance).  

\subsection{Model interpretation}

In order to probe how the XGB and MLP training parameterization describes the dead fuel-moisture values, we employed several methods to help explain the predicted outputs of the XGB and MLP models in terms of the input predictors. The permutation method measures the importance of each feature by computing the error difference before and after perturbing the feature value given some input $x$. Small changes in error typically indicate lower importance, and vice versa. 

Secondly, the SHapley Additive exPlanations (SHAP) \cite{shapley_book, lundberg2017} method seeks to quantify how much each feature value contributed to a model prediction, relative to the average prediction, in the feature's respective units. For example, a SHAP value for the input land surface temperature of +1 K, relative to an average value, means that the input explained that much of the predicted output value. As this example illustrates, there will be a SHAP value for each input that can be used to explain the output of either the XGB or MLP model. Formally, the SHAP value for a feature is its average marginal contribution as computed across all possible subsets of the model inputs that contain it \cite{shapley_book}.

Finally, applied only to the XGB model is the gain method for estimating importance. The gain estimates the relative feature importance according to how much the feature tree contributed to the total model FMC prediction. The larger the contribution the more important a feature, and vice versa.

\section{Results}

\subsection{Predictor group importance}

\begin{figure}[t!]
    \centering
    \includegraphics[width=\columnwidth]{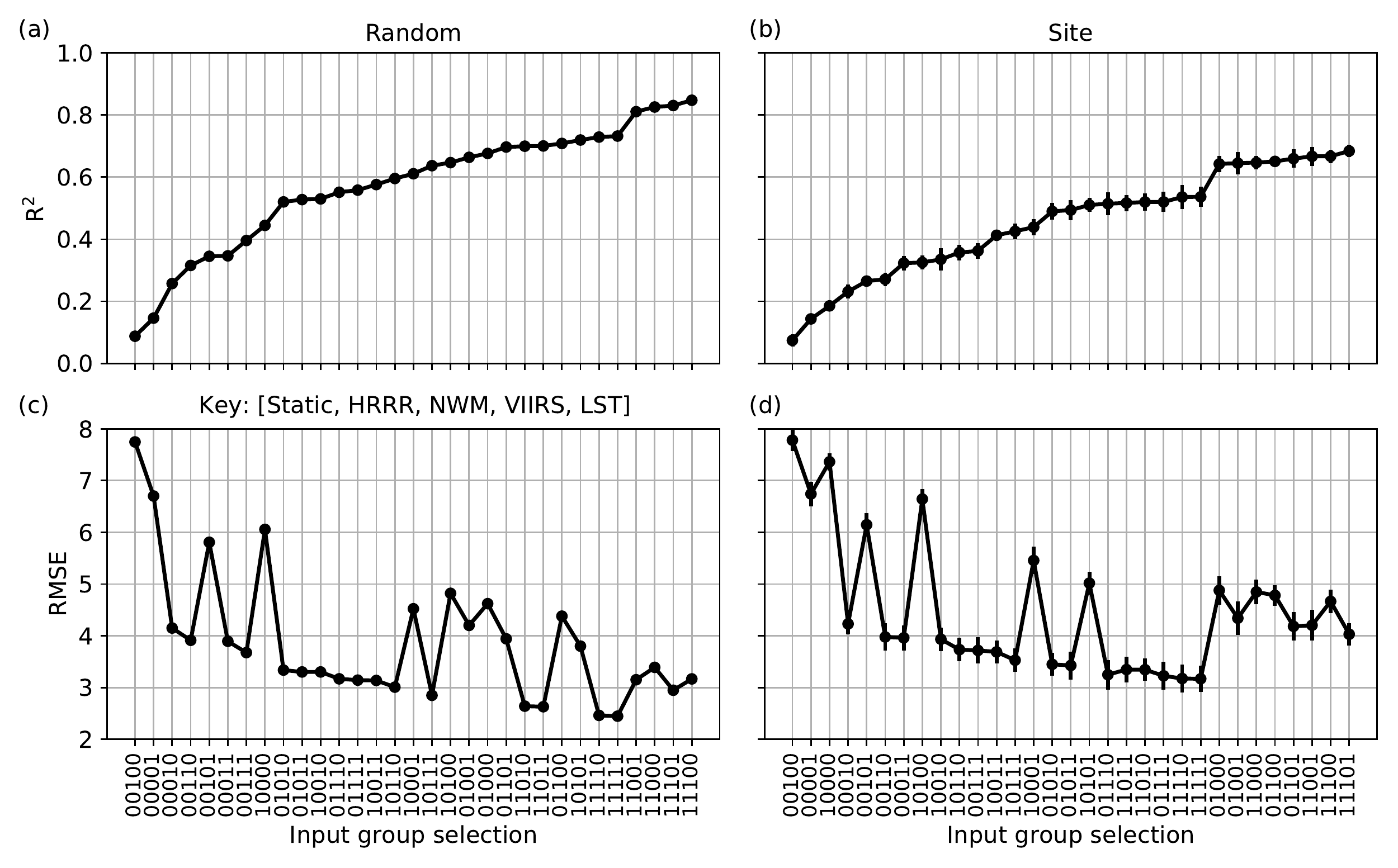}
    \caption{The XGB model was trained on 31 relevant combinations of the 5 possible input groups, using the same training and model hyperparameters (listed in appendix A1). The key indicates whether a group of input features was used during training, indicated by 1, and 0 if it was not. Panels (a) and (b) show the average $R^2$ score, while (c) and (d) show the average RMSE. The panels in (a) and (c) used the random validation split while (b) and (d) used the site validation split. For each split, the data points for both $R^2$ and RMSE were sorted using the $R^2$ score.}
    \label{fig:groups}
\end{figure}

We first determined which groups were the most important to use as potential predictors by training and testing the XGB model on all combinations of the 4 groups listed in Table \ref{table:predictors}, plus the LST treated as its own group (hence 31 total relevant combinations), on the 375-m data sets. The same model hyperparameters were used. Figure \ref{fig:groups} shows the performance metrics RMSE (bottom row) and $R^2$ (top row) for the random (left column) and site (right column) testing data splits. Both quantities were ordered by the computed $R^2$ (least to greatest). 

In the RMSE figures for both split types, the zig-zag pattern depends on the VIIRS surface reflectances set. When the reflectances are used as predictors, a clear drop in the RMSE is observed (e.g. improved performance), but there is not a similarly strong dependence in the computed $R^2$. This behavior is observed with both the MLP and LG models (not shown). The LST feature is also helpful but not necessary, as the performance gains over not including it are relatively small for XGB. The importance of VIIRS reflectances as a group is due to the relatively fast equilibration times between the dead fuels (mostly sticks and brush at 10-h) and the atmosphere, which are best captured by the twice-a-day retrievals. In contrast, the NWM variables do not seem to be necessary, as they hardly affect performance when used as predictors, which is reasonable because the 10-h fuel equilibrates with the atmosphere and not the soil.

In terms of lower RMSEs, the HRRR and VIIRS variables play the most important roles as predictor groups, and the (11111) models had the lowest ensemble average RMSE for both splits, with (01111) coming in a close second. Thus, the static predictors do not play a significant role in either random or site split routine. This is potentially useful for using the model outside of the CONUS (e.g., in Alaska) because the static variables are site-dependent and can be ignored as model inputs. However, this remains to be tested until Alaska data become available. Finally, the worst model in both split cases is that utilizing the NWM only.

\subsection{Model performance}

\setlength{\extrarowheight}{8pt}
\begin{table}[t!]
\begin{center}
\begin{tabular}{ l | c | c | c }
RMSE (Split) & Linear Regression & XGB & MLP \\
\hline
Random 375 m (11111) & 3.44 $\pm$ 0.01  & 2.48 $\pm$ 0.02 & 2.37 $\pm$ 0.08 \\ 
Random 2250 m (11111) & 3.65 $\pm$ 0.01 & 2.64 $\pm$ 0.01 & 2.29 $\pm$ 0.17 \\
Site 375 m (01111) & 4.12 $\pm$ 0.01 & 3.56 $\pm$ 0.09 & 3.71 $\pm$ 0.02 \\ 
Site 2250 m (01111) & 3.53 $\pm$ 0.01 & 3.04 $\pm$ 0.04 & 3.21 $\pm$ 0.02  \\ 
\hline
R$^2$ (Split) &  &  &  \\
\hline
Random 375 m (11111) & 0.45 $\pm$ 0.001 & 0.72 $\pm$ 0.01 & 0.74 $\pm$ 0.02 \\
Random 2250 m (11111) & 0.45 $\pm$ 0.002 & 0.71 $\pm$ 0.01 & 0.78 $\pm$ 0.03 \\
Site 375 m (01111) & 0.37 $\pm$ 0.003 & 0.53 $\pm$ 0.02 & 0.49 $\pm$ 0.01 \\ 
Site 2250 m (01111) & 0.51 $\pm$ 0.001 & 0.64 $\pm$ 0.01 & 0.59 $\pm$ 0.01 \\ 
\hline\hline
\end{tabular}
\end{center}
\caption{The three ML models are compared using the RMSE and R2 metrics. The rows show the metric values for the two splits (Random and Site) at 375-m and 2250-m resolutions. In all cases, a 10-fold cross-validation splitting routine was used to estimate the mean and standard deviation of each quantity for the testing data set. } %
\label{results:metrics}
\end{table}

Table~\ref{results:metrics} compares the computed bulk performance metrics for the LG model and hyperparameter optimized XGB and MLP models, for the random and site test splits, respectively. Models trained on the random split used all predictors while those trained on the site split left out the static predictors. Overall, the models always perform better on the random split relative to the site split, which is expected as the random splitting most likely contains correlated subgroups of data in both training and testing splits. The performance of models trained on the site split, which represent the more realistic performance we might expect when the model is in operation over both CONUS and Alaska, is always lower by comparison.  Additionally, slightly better performance is usually observed when models are trained on the 2250-m relative to the 375-m data sets (but this is not always the case for XGB). 

Table~\ref{results:metrics} also shows clearly that both XGB and the MLP models outperform the LG on all metrics and on both data splits at 375-m and 2250-m resolutions. The MLP is observed to outperform the XGB on the random split, while XGB outperforms or is comparable in performance to the MLP on the site split, which did not include the static (site dependent) variables as predictors. This is important because operationally XGB is much faster to use compared to the relatively large MLP model, which contains 6 hidden layers each of size 6,427 neurons.

\begin{figure}[t!]
    \centering
    \includegraphics[width=\columnwidth]{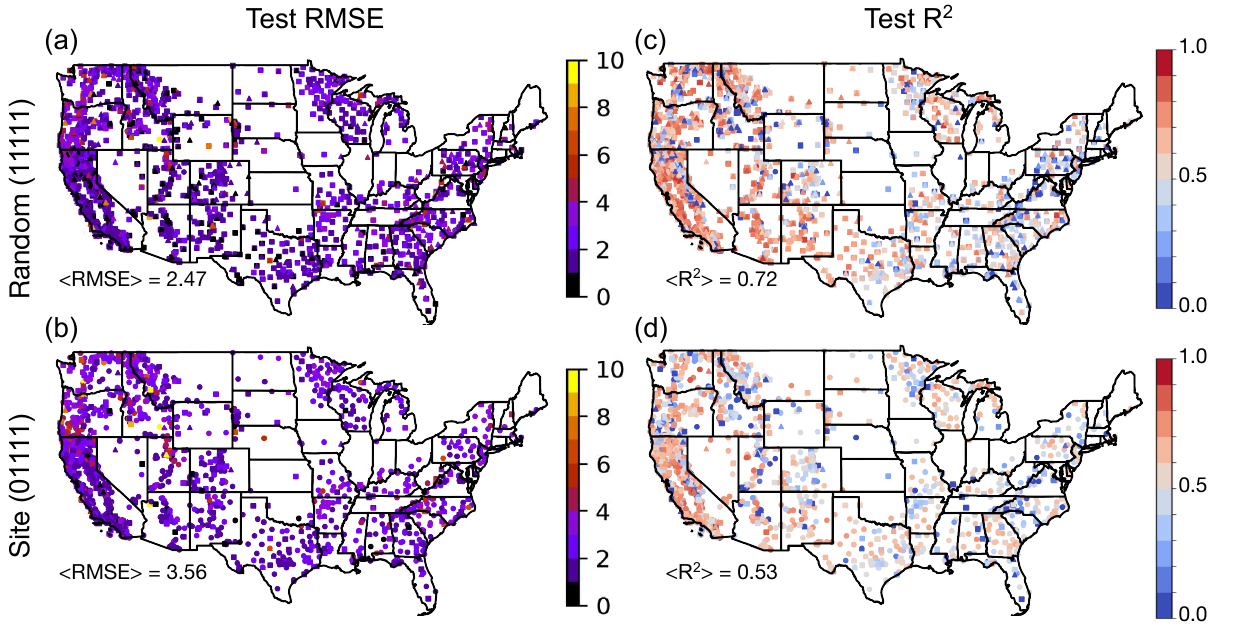}
    \caption{The RMSE and $R^2$ metrics are shown in the two columns, respectively, for XGB models trained on the random and site data splits (top and bottom rows, respectively). In all panels, circles show training points, squares show validation points, and triangles show test data points. In (c) and (d), points with negative R$^2$ values are clipped to zero.}
    \label{fig:conus_metrics}
\end{figure}

Figure~\ref{fig:conus_metrics} shows the performance metrics computed at the site level over CONUS for the XGB model trained on the 375-m site data split. Clearly, the model produces the best results in drier areas (for example, the desert southwest and southern California, which has the lowest RMSE and highest $R^2$ scores) where the FMC does not change as much. On the other hand, the performance varies more in the Pacific Northwest where more variability in fuel moisture is occurring, which is logically harder for the model to capture as well. The model trained on the random split shows better performance as expected, though both models show similar trends (such as better performance in drier areas). 

Overall, the RMSE values look ``good'', the model is performing well relative to other FMC retrievals over CONUS. With either ML model and the current 3-year data set, we still need to know when a trained model performs better compared to climatographical baselines for it to be effectively useful, otherwise the climatography estimates should be used. Note that the relatively low $R^2$ score for the model trained on the site split indicates that further performance improvements need to be sought after, but the climatography scores tell when the model is practically useful.

\subsection{Model skill}

\begin{figure}[t!]
    \centering
    \includegraphics[width=\columnwidth]{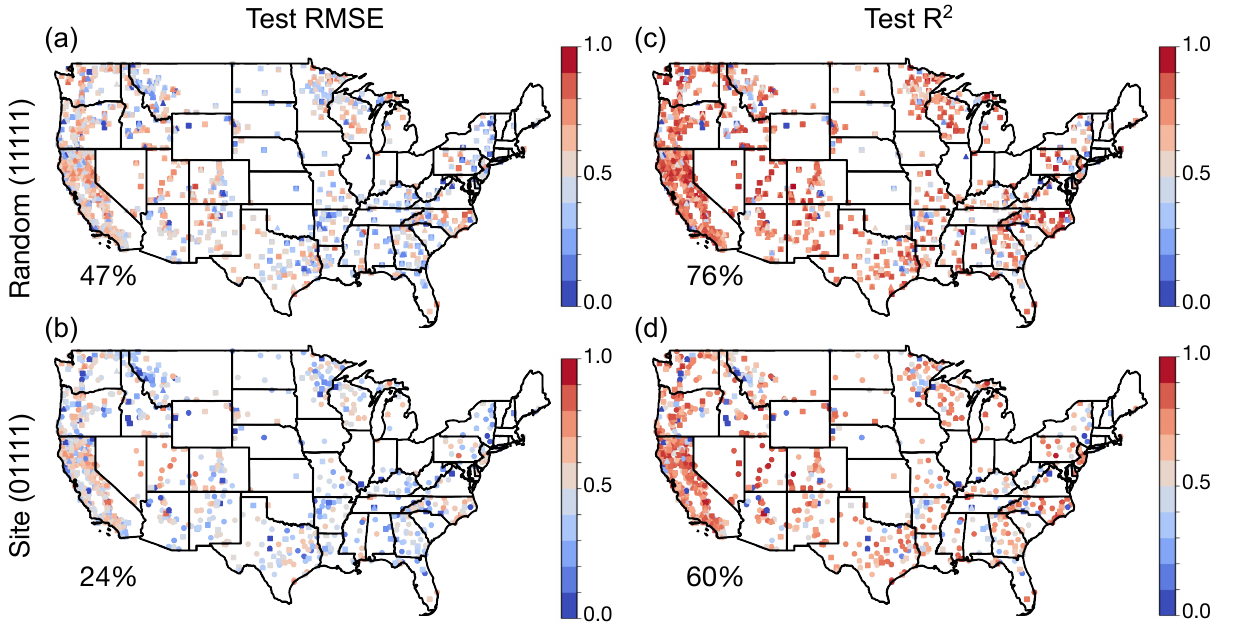}
    \caption{The computed hourly skill scores for RMSE and $R^2$ are shown in the left and right columns, respectively. The top and bottom rows show the random and site splits, respectively. In all panels, circles show training data, squares show validation data, and triangles show test data.}
    \label{fig:conus_clim_hour}
\end{figure}

\begin{figure}[t!]
    \centering
    \includegraphics[width=\columnwidth]{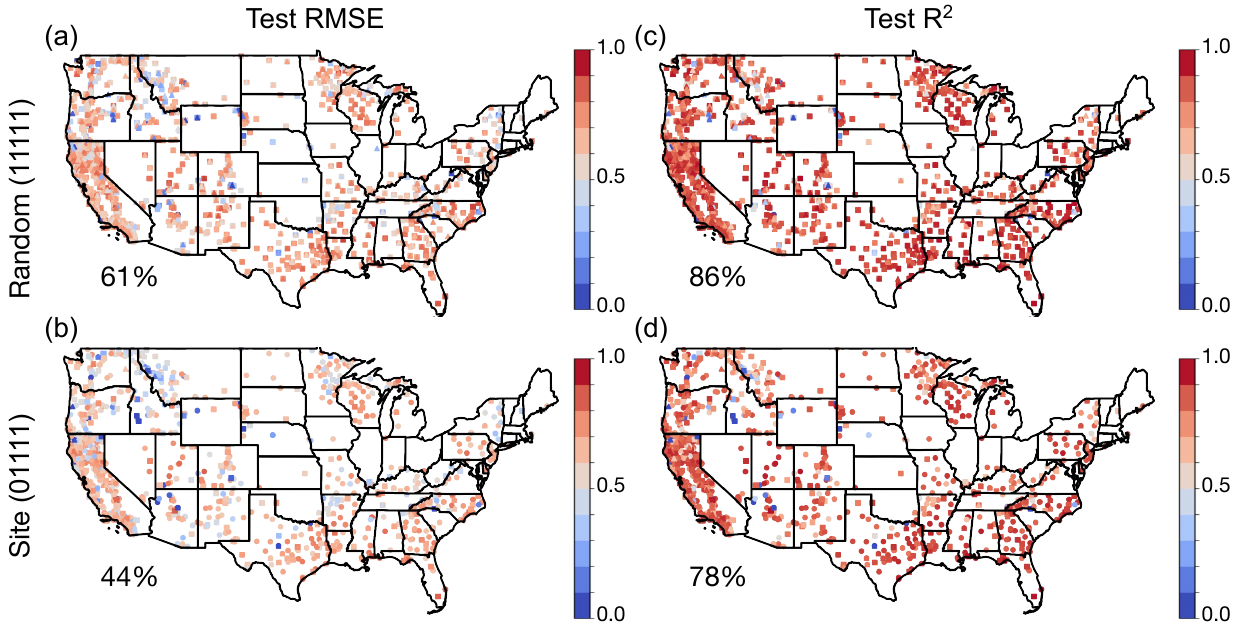}
    \caption{The computed daily skill scores for RMSE and $R^2$ are shown in the left and right columns, respectively. The top and bottom rows show the random and site splits, respectively. In all panels, circles show training data, squares show validation data, and triangles show test data.}
    \label{fig:conus_clim_day}
\end{figure}

Figures~\ref{fig:conus_clim_hour} and \ref{fig:conus_clim_day} show the spatial distributions of the model skill scores computed using Equations \ref{skill_rmse} and \ref{skill_r2} at the site level across time, using hourly and daily climatography estimates, respectively. Both the figures show that the models outperform climatography estimates for most of the sites on the CONUS map, with fairly consistent skill observed across different regions according to both RMSE and $R^2$. The outlier sites, where the climatography estimate has the higher skill, tend to be in the mountain west. Overall, the random and site RMSE performances improved over the daily climatography estimates by 61\% and 44\%, respectively. The hourly climatography is a better model than the daily climatography, and the skill scores are, as expected, smaller, with an improvement of +47\% and +24\% for the random and site splits, respectively. Similar increases are seen with the $R^2$ skill score metric. The figures also show that the training, validation, and testing splits yielded similar results, indicating the model is not overfitted to the training data split in each case. 

\begin{figure}[t!]
    \centering
    \includegraphics[width=\columnwidth]{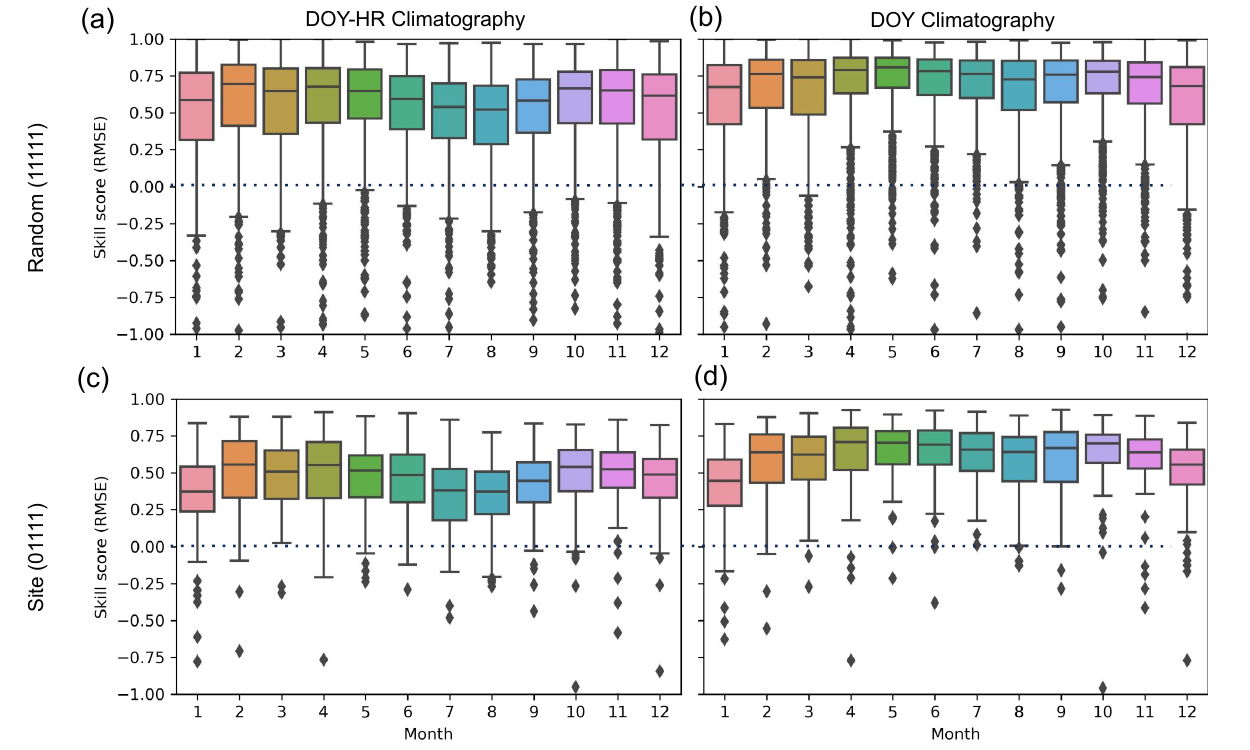}
    \caption{The hour-of-the-day, day-of-the-year (DOY-HR) and day-of-the-year (DOY) skill scores for RMSE are shown in the left and right columns, respectively. The top and bottom rows show the random and site splits, respectively. Points above the dashed line in the panels indicate the model is more skillful relative to climatography estimates, while points below indicate that climatography estimates are more skillful. Points with negative values are clipped to zero. In all panels, circles show training data, squares show validation data, and triangles show test data.}
    \label{fig:clim_hour}
\end{figure}

\begin{figure}[t!]
    \centering
    \includegraphics[width=\columnwidth]{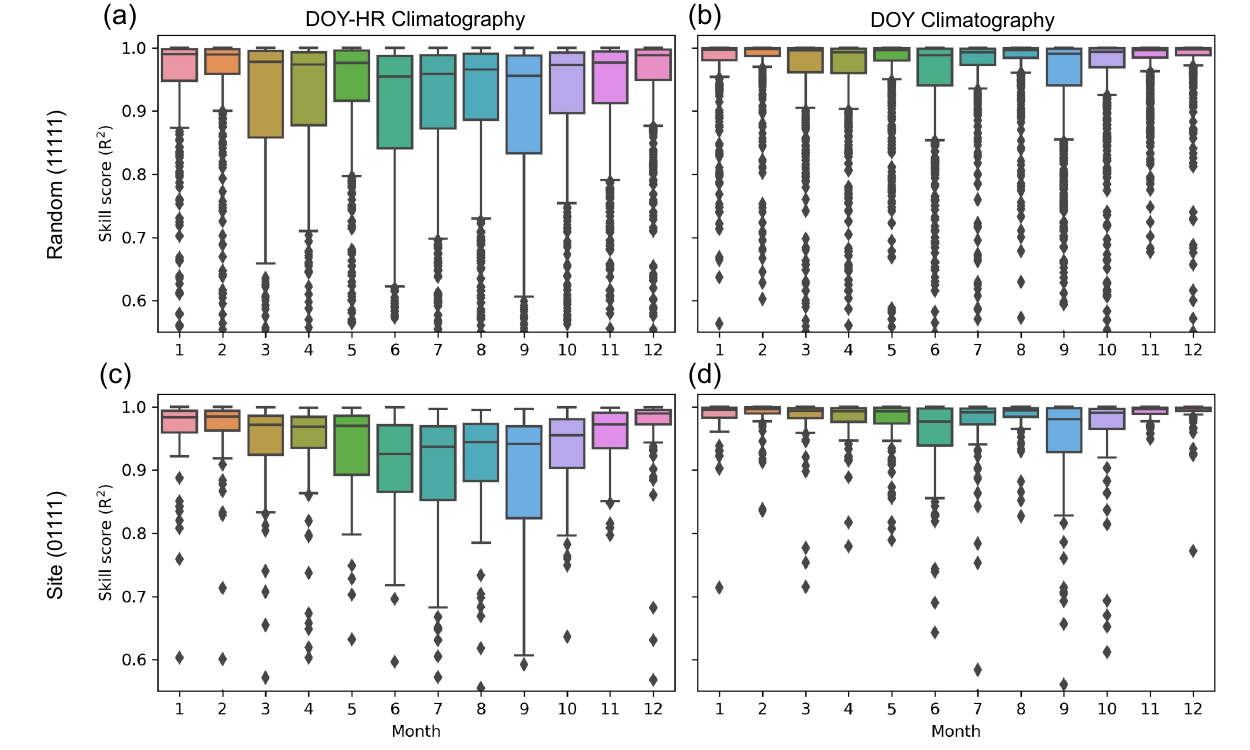}
    \caption{The hour-of-the-day, day-of-the-year (DOY-HR) and day-of-the-year (DOY) skill scores for $R^2$ are shown in the left and right columns, respectively. The top and bottom rows show the random and site splits, respectively. Points with negative values are clipped to zero. In all panels, circles show training data, squares show validation data, and triangles show test data.}
    \label{fig:clim_day}
\end{figure}

In order to understand the time-dependence of the model performance, Figures~\ref{fig:clim_hour} and \ref{fig:clim_day} plot the RMSE and $R^2$ skill scores for the random and site test splits by month, respectively, starting with January (1) and ending with December (12). The data points were computed by averaging over hour-of-the-day, day-of-the-year (DOY-HR) and day-of-the-year(DOY), and then grouped by month to create the box-whisker diagrams in the figures. 

Figures~\ref{fig:clim_hour} shows for both splits and both climatography estimates, that the models are more skillful throughout the year with more than 75\% of points having skill larger than 0 at all times. However, there is still a small fraction of model predictions that are worse compared to the climatography estimates, and all outliers are less skillful than climatography estimates of DOY-HR. By contrast, the results are improved for DOY compared to DOY-HR. 

There is also a clear seasonal performance dependence when compared to DOY-HR climatography. In particular, the model RMSE performance peaks in the spring and fall, and bottoms out in the summer and winter (Figure~\ref{fig:clim_hour}(a) and (c)). Model R$^2$ performance against DOY also shows performance hitting a minimum in the winter time, while remaining mainly flat during the other three seasons (Figure~\ref{fig:clim_day}(a) and (c)). The lower model skill on the hourly climotagraphy indicates, as expected, that hourly climatography is more skillful than daily. At the site level, the model has no information about the sites, but climatography estimates require site-level data, so this is not really a fair comparison for the models trained on the site split. However, it should be a more fair comparison for the models trained on the random split.

\subsection{Predictor importance}

\begin{figure*}[ht!]
    \centering
    \includegraphics[width=0.8\columnwidth]{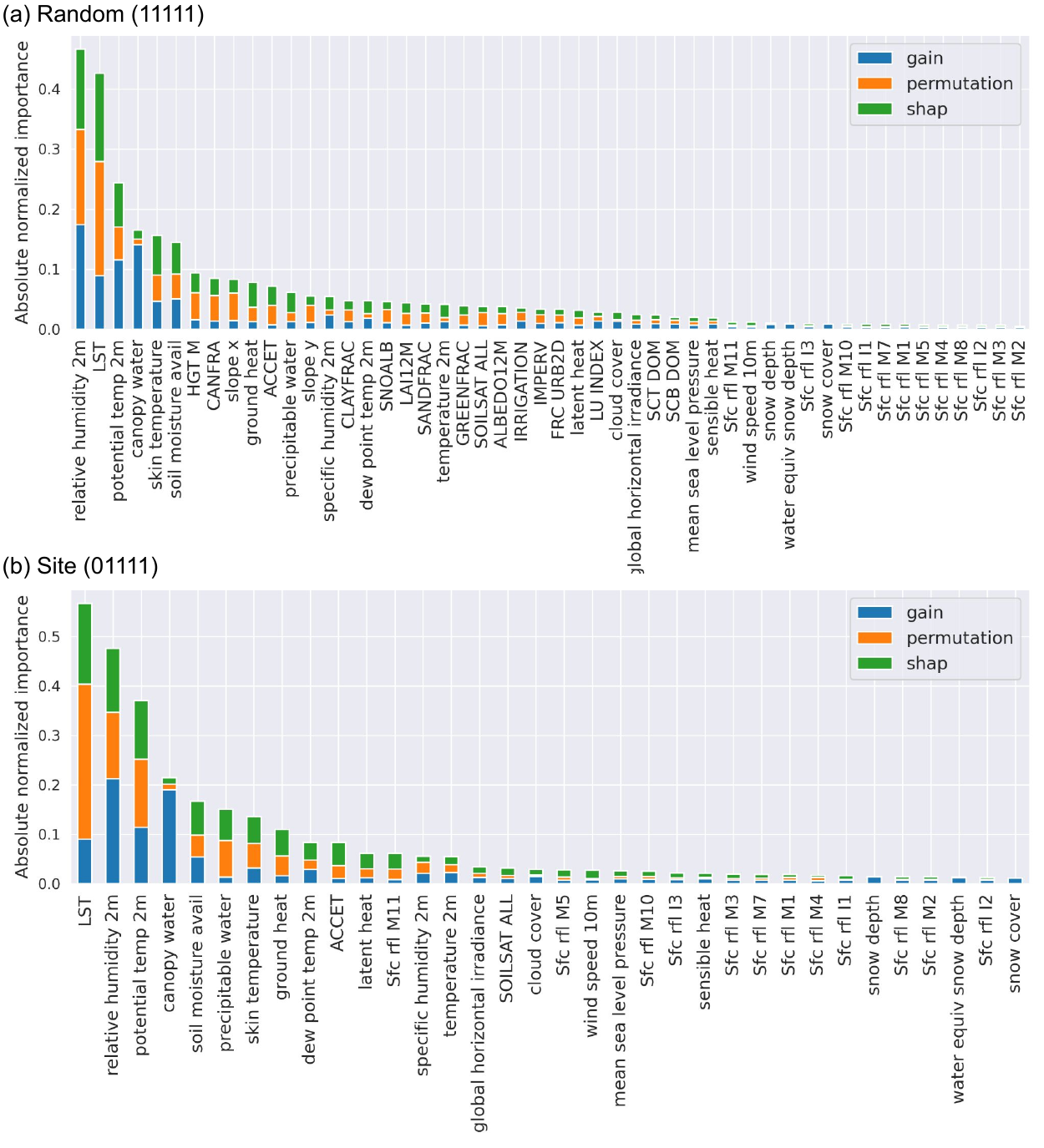}
    \caption{The computed predictor importance's are shown stacked one on top the other for the gain, permutation, and SHAP metrics. Panels (a) and (b) show the XGB model trained on the random split (using all predictors, e.g., (11111)), and site model using all predictors except the static group (e.g., (01111), respectively.}
    \label{fig:explain}
\end{figure*}

Finally, we quantify the relative predictor importance using the permutation, SHAP, and gain methods. Figure~\ref{fig:explain}(a) and (b) shows the three computed quantities for each predictor used in random and site test splits, respectively. The figures show the three importance metrics sorted from greatest to least importance after being summed. 

Both models predict LST medium and relative humidity at 2 m as the top two predictors (in different order). Additionally, the top six of seven predictors are the same for the two models, which are LST, relative humidity at 2 m, potential temperature at 2 m, canopy water, soil moisture availability, and skin temperature. In other words, the FMC predictions are mostly explained using temperature and moisture predictors, which seems physically reasonable. The relative importance of the top five predictors also dominates those in the bottom half. However, the three methods do not rank the predictors identically. For example, the gain approach clearly differs from the other two on the importance of canopy water -- only it shows that predictor being relatively significant. By contrast, the SHAP and permutation approaches suggest precipitable water has higher importance.

The importance of the VIIRS reflectances are also absent in both figures, and are always in the bottom half of predictor importance. This apparent lack of importance quantification by all three methods is due to the high correlation among the reflectances. When combined with the results from Figure \ref{fig:groups}, the VIIRS reflectances are understood to be important as a group, but any one individual band alone is not sufficient to contribute to the explanation of FMC, as Figure~\ref{fig:explain} shows. 

\section{Discussion}

Overall, the above analyses highlight several important data and model choices, as well as deployment considerations involved in modeling FMC with machine learning. First, the performance of ML models is highly dependent on the data sources selected as fuel-moisture predictors. Clearly, the most important predictor groups needed to produce skillful XGB (and MLP), relative to measured climatographies, are the HRRR and VIIRS retrievals. The VIIRS retrievals contribute as a group due to high band-correlation, while a small number of individual predictors in the HRRR group have relatively high importance according to the explainability techniques used. In Figure~\ref{fig:explain}, the LST predictor has high overall explainable importance, but it has low importance in Figure~\ref{fig:groups} when included as an input group. Thus, the predictor can be removed without causing a significant performance decline. This is primarily due to its strong correlation with the potential temperature at 2-m in the HRRR group. Recall that when both HRRR and VIIRS retrievals are not used as model inputs (e.g., the surface temperature predictors are removed) the RMSE performance drops significantly (Figure~\ref{fig:groups}), especially for the site split, corroborating the high importance of the surface temperature predictors (Figure~\ref{fig:explain}). We also observed essentially the same group importance result for the MLP model (the results are not shown but are closely comparable to that presented for XGB). The overall importance of the two groups corroborate the dynamic relationship between the 10-h fuel and the atmosphere, and with soil moisture. 

Next, the random and site approaches to splitting the data set before training ML models demonstrate the difference between an ideal scenario and expected performance when carefully preparing training and validation splits. Generally, the ML model performs worse on the site split, while the random split performs better due to the high space and time correlation between the training splits. The site split approach has an added advantage since the model validation does not depend on specific sites, unlike the random split approach. Therefore, ML models trained on decorrelated data splits may be more useful in regions outside of the training data, such as Alaska, but still with similar climates and geographies. It is worth noting that both splits are not overfitted on the hold-out splits, as predicted FMC distributions and testing metrics look similar across the training, validation, and testing splits. Spatially, the performance is relatively uniform over CONUS, with likely some California bias. As more data become available, these problems can be resolved. The models, including new variations, will be tested with the additional data, although for early release, we intend to use ML models trained on the site split.

Finally, even though we primarily focused on the performance of the XGB model, the best MLPs perform similarly on the site splits and outperform XGB on the random splits. Therefore, which model should be used in deployment? For several reasons, the XGB model will be used initially. First, the optimized architectures contain millions of fitted parameters, which necessitates GPU computation during inference to obtain the best performance, as well as future training when more data is available. By comparison, even the optimized XGB found here (which are pretty large), can be orders of magnitude faster to run relative to the MLP on the GPU (depending on the overall size of the MLP). Secondly, the MLPs are able to obtain the reported performance, in part due to the transformation of the predictors and the predicted FMC into z-scores before training the model, which requires performing the inverse transformation on the predicted FMC during inference. Even though the same transformations are applied to the XGB model (and the linear regression baseline), this is not usually required for the XGB model. Applying preprocessing transformations may limit the ability of trained models to be generalized beyond the distributions present in the training data sets, and furthermore represents another computational step needed during deployment. Lastly, while we did not study their performance here, the XGB model can be trained on data sets with missing values, whereas neural networks currently require masking or imputation strategies. 

\section{Conclusion}

In summary, we have found that hyper-parameter optimized XGB and fully-connected, feed forward neural architectures are both skillful with respect to hourly and daily climatographies for the task of 10-h dead FMC prediction. By exploring all combinations of input groups, the main predictor sources are found to be from HRRR and VIIRS. In particular, the most important HRRR predictors are those relating to temperature near the surface and available moisture. By contrast, the VIIRS bands seems to be important only when included as a group rather than any one band dominating either models' explanations. This latter finding is the result of fast-equilibration of 10-h dead sources and the atmosphere and soil. Even though the models provide better estimates relative to known climatological estimates, there are still performance gains that may be realized by including more data sources. This and the prediction of live FMC are the main objectives for future studies. 

\acknowledgments{We would like to acknowledge high-performance computing support from Cheyenne and Casper \cite{Cheyenne} provided by NCAR's Computational and Information Systems Laboratory, sponsored by the National Science Foundation. JSS would like to thank Thomas Martin (Unidata) for a careful reading of the manuscript and for providing helpful comments. The neural networks described here and simulation code used to train and test the models are archived at \url{https://github.com/NCAR/fmc_viirs}.  This research was funded by JPSS grant number R4310383.}

\bibliographystyle{unsrt}
\bibliography{references}

\begin{thebibliography}{10}

\bibitem{CBO}
{Congressional Budget Office}.
\newblock {CBO Publication 57970: Wildfires}.
\newblock Technical report, {US Congress}, 2022.

\bibitem{wiki:marshall}
{Wikipedia contributors}.
\newblock Marshall fire --- {W}ikipedia{,} the free encyclopedia, 2021.
\newblock [Online; accessed 7-February-2023].

\bibitem{coen2013wrf}
Janice~L Coen, Marques Cameron, John Michalakes, Edward~G Patton, Philip~J
  Riggan, and Kara~M Yedinak.
\newblock Wrf-fire: coupled weather--wildland fire modeling with the weather
  research and forecasting model.
\newblock {\em Journal of Applied Meteorology and Climatology}, 52(1):16--38,
  2013.

\bibitem{rothermel1972}
R.C. Rothermel.
\newblock {\em A Mathematical Model for Predicting Fire Spread in Wildland
  Fuels}.
\newblock USDA Forest Service research paper INT. Intermountain Forest \& Range
  Experiment Station, Forest Service, U.S. Department of Agriculture, 1972.

\bibitem{yebra2013global}
Marta Yebra, Philip~E Dennison, Emilio Chuvieco, David Ria{\~n}o, Philip
  Zylstra, E~Raymond Hunt~Jr, F~Mark Danson, Yi~Qi, and Sara Jurdao.
\newblock A global review of remote sensing of live fuel moisture content for
  fire danger assessment: Moving towards operational products.
\newblock {\em Remote Sensing of Environment}, 136:455--468, 2013.

\bibitem{wfas}
{US Forest Service}.
\newblock Dead fuel moisture - nfdrs.

\bibitem{chuvieco2004conversion}
Emilio Chuvieco, Inmaculada Aguado, and Alexandros~P Dimitrakopoulos.
\newblock Conversion of fuel moisture content values to ignition potential for
  integrated fire danger assessment.
\newblock {\em Canadian Journal of Forest Research}, 34(11):2284--2293, 2004.

\bibitem{aguado2007estimation}
I~Aguado, E~Chuvieco, R~Bor{\'e}n, and H~Nieto.
\newblock Estimation of dead fuel moisture content from meteorological data in
  mediterranean areas. applications in fire danger assessment.
\newblock {\em International Journal of Wildland Fire}, 16(4):390--397, 2007.

\bibitem{nolan2016predicting}
Rachael~H Nolan, V{\'\i}ctor~Resco de~Dios, Matthias~M Boer, Gabriele Caccamo,
  Michael~L Goulden, and Ross~A Bradstock.
\newblock Predicting dead fine fuel moisture at regional scales using vapour
  pressure deficit from modis and gridded weather data.
\newblock {\em Remote Sensing of Environment}, 174:100--108, 2016.

\bibitem{nolan2016large}
Rachael~H Nolan, Matthias~M Boer, Victor Resco~de Dios, Gabriele Caccamo, and
  Ross~A Bradstock.
\newblock Large-scale, dynamic transformations in fuel moisture drive wildfire
  activity across southeastern australia.
\newblock {\em Geophysical Research Letters}, 43(9):4229--4238, 2016.

\bibitem{boer2017changing}
Matthias~M Boer, Rachael~H Nolan, V{\'\i}ctor Resco De~Dios, Hamish Clarke,
  Owen~F Price, and Ross~A Bradstock.
\newblock Changing weather extremes call for early warning of potential for
  catastrophic fire.
\newblock {\em Earth's Future}, 5(12):1196--1202, 2017.

\bibitem{hiers2019fine}
J~Kevin Hiers, Christina~L Stauhammer, Joseph~J O’Brien, Henry~L Gholz,
  Timothy~A Martin, John Hom, and Gregory Starr.
\newblock Fine dead fuel moisture shows complex lagged responses to
  environmental conditions in a saw palmetto (serenoa repens) flatwoods.
\newblock {\em Agricultural and Forest Meteorology}, 266:20--28, 2019.

\bibitem{lee2020estimation}
HoonTaek Lee, Myoungsoo Won, Sukhee Yoon, and Keunchang Jang.
\newblock Estimation of 10-hour fuel moisture content using meteorological
  data: a model inter-comparison study.
\newblock {\em Forests}, 11(9):982, 2020.

\bibitem{cawson2020corrigendum}
Jane~G Cawson, Petter Nyman, Christian Schunk, Gary~J Sheridan, Thomas~J Duff,
  Kelsy Gibos, William~D Bovill, Marco Conedera, Gianni~B Pezzatti, and Annette
  Menzel.
\newblock Corrigendum to: Estimation of surface dead fine fuel moisture using
  automated fuel moisture sticks across a range of forests worldwide.
\newblock {\em International Journal of Wildland Fire}, 29(6):560--560, 2020.

\bibitem{masinda2021prediction}
Maombi~Mbusa Masinda, Fei Li, Qi~Liu, Long Sun, and Tongxin Hu.
\newblock Prediction model of moisture content of dead fine fuel in forest
  plantations on maoer mountain, northeast china.
\newblock {\em Journal of Forestry Research}, 32(5):2023--2035, 2021.

\bibitem{rs13214224}
Eleni Dragozi, Theodore~M. Giannaros, Vasiliki Kotroni, Konstantinos
  Lagouvardos, and Ioannis Koletsis.
\newblock Dead fuel moisture content (dfmc) estimation using modis and
  meteorological data: The case of greece.
\newblock {\em Remote Sensing}, 13(21), 2021.

\bibitem{marino2020investigating}
Eva Marino, Marta Yebra, Mariluz Guill{\'e}n-Climent, Nur Algeet, Jos{\'e}~Luis
  Tom{\'e}, Javier Madrigal, Mercedes Guijarro, and Carmen Hernando.
\newblock Investigating live fuel moisture content estimation in fire-prone
  shrubland from remote sensing using empirical modelling and rtm simulations.
\newblock {\em Remote Sensing}, 12(14):2251, 2020.

\bibitem{caccamo2011monitoring}
G~Caccamo, LA~Chisholm, RA~Bradstock, Marjetta~L Puotinen, and BG~Pippen.
\newblock Monitoring live fuel moisture content of heathland, shrubland and
  sclerophyll forest in south-eastern australia using modis data.
\newblock {\em International Journal of Wildland Fire}, 21(3):257--269, 2011.

\bibitem{stow2006time}
Douglas Stow, Madhura Niphadkar, and John Kaiser.
\newblock Time series of chaparral live fuel moisture maps derived from modis
  satellite data.
\newblock {\em International Journal of Wildland Fire}, 15(3):347--360, 2006.

\bibitem{peterson2008mapping}
Seth~H Peterson, Dar~A Roberts, and Philip~E Dennison.
\newblock Mapping live fuel moisture with modis data: A multiple regression
  approach.
\newblock {\em Remote Sensing of Environment}, 112(12):4272--4284, 2008.

\bibitem{NIETO2010861}
Héctor Nieto, Inmaculada Aguado, Emilio Chuvieco, and Inge Sandholt.
\newblock Dead fuel moisture estimation with msg–seviri data. retrieval of
  meteorological data for the calculation of the equilibrium moisture content.
\newblock {\em Agricultural and Forest Meteorology}, 150(7):861--870, 2010.

\bibitem{zormpas2017dead}
Konstantinos Zormpas, Christos Vasilakos, Nikos Athanasis, Nikos Soulakellis,
  and Kostas Kalabokidis.
\newblock Dead fuel moisture content estimation using remote sensing.
\newblock {\em European Journal of Geography}, 8(5):17--32, 2017.

\bibitem{McCandless2020}
Tyler~C McCandless, Branko Kosovic, and William Petzke.
\newblock Enhancing wildfire spread modelling by building a gridded fuel
  moisture content product with machine learning.
\newblock {\em Machine Learning: Science and Technology}, 1(3):035010, aug
  2020.

\bibitem{fan2021physics}
Chunquan Fan and Binbin He.
\newblock A physics-guided deep learning model for 10-h dead fuel moisture
  content estimation.
\newblock {\em Forests}, 12(7):933, 2021.

\bibitem{SHMUEL2022119897}
Assaf Shmuel, Yiftach Ziv, and Eyal Heifetz.
\newblock Machine-learning-based evaluation of the time-lagged effect of
  meteorological factors on 10-hour dead fuel moisture content.
\newblock {\em Forest Ecology and Management}, 505:119897, 2022.

\bibitem{xie2022retrieval}
Jiangjian Xie, Tao Qi, Wanjun Hu, Huaguo Huang, Beibei Chen, and Junguo Zhang.
\newblock Retrieval of live fuel moisture content based on multi-source remote
  sensing data and ensemble deep learning model.
\newblock {\em Remote Sensing}, 14(17):4378, 2022.

\bibitem{capps2021modelling}
Scott~B Capps, Wei Zhuang, Rui Liu, Tom Rolinski, and Xin Qu.
\newblock Modelling chamise fuel moisture content across california: A machine
  learning approach.
\newblock {\em International Journal of Wildland Fire}, 31(2):136--148, 2021.

\bibitem{zhu2021live}
Liujun Zhu, Geoffrey~I Webb, Marta Yebra, Gianluca Scortechini, Lynn Miller,
  and Fran{\c{c}}ois Petitjean.
\newblock Live fuel moisture content estimation from modis: A deep learning
  approach.
\newblock {\em ISPRS Journal of Photogrammetry and Remote Sensing}, 179:81--91,
  2021.

\bibitem{xgboost}
Tianqi Chen and Carlos Guestrin.
\newblock Xgboost: {A} scalable tree boosting system.
\newblock In Balaji Krishnapuram, Mohak Shah, Alexander~J. Smola, Charu~C.
  Aggarwal, Dou Shen, and Rajeev Rastogi, editors, {\em Proceedings of the 22nd
  {ACM} {SIGKDD} International Conference on Knowledge Discovery and Data
  Mining, San Francisco, CA, USA, August 13-17, 2016}, pages 785--794. {ACM},
  2016.

\bibitem{mlp_model}
Yury Gorishniy, Ivan Rubachev, Valentin Khrulkov, and Artem Babenko.
\newblock Revisiting deep learning models for tabular data.
\newblock In Marc'Aurelio Ranzato, Alina Beygelzimer, Yann~N. Dauphin, Percy
  Liang, and Jennifer~Wortman Vaughan, editors, {\em Advances in Neural
  Information Processing Systems 34: Annual Conference on Neural Information
  Processing Systems 2021, NeurIPS 2021, December 6-14, 2021, virtual}, pages
  18932--18943, 2021.

\bibitem{kossen2021self}
Jannik Kossen, Neil Band, Clare Lyle, Aidan~N Gomez, Thomas Rainforth, and
  Yarin Gal.
\newblock Self-attention between datapoints: Going beyond individual
  input-output pairs in deep learning.
\newblock {\em Advances in Neural Information Processing Systems},
  34:28742--28756, 2021.

\bibitem{tabular_xgb1}
Ravid Shwartz{-}Ziv and Amitai Armon.
\newblock Tabular data: Deep learning is not all you need.
\newblock {\em Inf. Fusion}, 81:84--90, 2022.

\bibitem{tabular_xgb2}
L{\'{e}}o Grinsztajn, Edouard Oyallon, and Ga{\"{e}}l Varoquaux.
\newblock Why do tree-based models still outperform deep learning on tabular
  data?
\newblock {\em CoRR}, abs/2207.08815, 2022.

\bibitem{rumelhart1986}
David~E Rumelhart, Geoffrey~E Hinton, and Ronald~J Williams.
\newblock Learning representations by back-propagating errors.
\newblock {\em Nature}, 323(6088):533--536, 1986.

\bibitem{bergstra2011algorithms}
James Bergstra, R{\'e}mi Bardenet, Yoshua Bengio, and Bal{\'a}zs K{\'e}gl.
\newblock Algorithms for hyper-parameter optimization.
\newblock {\em Advances in neural information processing systems}, 24, 2011.

\bibitem{shapley_book}
L.~S. Shapley.
\newblock {\em 17. A Value for n-Person Games}, pages 307--318.
\newblock Princeton University Press, 2016.

\bibitem{lundberg2017}
Scott~M Lundberg and Su-In Lee.
\newblock A unified approach to interpreting model predictions.
\newblock In {\em Proceedings of the 31st international conference on neural
  information processing systems}, pages 4768--4777, 2017.

\bibitem{Cheyenne}
{Computational and Information Systems Laboratory, CISL}.
\newblock {Cheyenne: HPE/SGI ICE XA System (NCAR Community Computing)}.
\newblock Technical report, {National Center for Atmospheric Research}, 2020.

\end{thebibliography}

\end{document}